\documentclass[11pt]{article}
\usepackage{geometry}
\usepackage{coling2020}
\usepackage{times}
\usepackage{url}
\usepackage{latexsym}
\usepackage{microtype}
\usepackage{tabularx}
\usepackage{graphicx}
\usepackage{adjustbox}
\usepackage{amsmath}
\usepackage{dirtytalk}
\usepackage{enumitem}
\usepackage{booktabs}
\usepackage{multirow}

\colingfinalcopy 

\title{gundapusunil at SemEval-2020 Task 8: Multimodal Memotion Analysis}

\author{Sunil Gundapu \\
  Language Technologies Research Centre \\
  KCIS, IIIT Hyderabad \\
  Telangana, India \\
  {\tt sunil.g@research.iiit.ac.in} \\\And
  Radhika Mamidi \\
  Language Technologies Research Centre \\
  KCIS, IIIT Hyderabad \\
  Telangana, India \\
  {\tt radhika.mamidi@iiit.ac.in} \\}

\date{}

\begin{document}
\maketitle
\begin{abstract}
  Recent technological advancements in the Internet and Social media usage have resulted in the evolution of faster and efficient platforms of communication. These platforms include visual, textual and speech mediums and have brought a unique social phenomenon called Internet memes. Internet memes are in the form of images with witty, catchy, or sarcastic text descriptions. In this paper, we present a multi-modal sentiment analysis system using deep neural networks combining Computer Vision and Natural Language Processing. Our aim is different than the normal sentiment analysis goal of predicting whether a text expresses positive or negative sentiment; instead, we aim to classify the Internet meme as a positive, negative, or neutral, identify the type of humor expressed and quantify the extent to which a particular effect is being expressed. Our system has been developed using CNN and LSTM and outperformed the baseline score.
\end{abstract}

\blfootnote{
    %
    %
     \hspace{-0.65cm}  
     This work is licensed under a Creative Commons 
     Attribution 4.0 International Licence.
     Licence details:
     \url{http://creativecommons.org/licenses/by/4.0/}.
    %
    %
}

\section{Introduction}

According to Wikipedia article on Internet Memes, \say{A meme is an idea, behavior, or style that spreads from person to person within a culture often with the aim of conveying a particular phenomenon, theme, or meaning represented by the meme}. 

Meme is not only about the humorous picture, the Internet culture, or the sentiment that passes along, but also about the richness and distinctiveness of its language: it is often greatly structured with unusual written style. Consider an example, the cat meme in Figure 1 often include superimposed text description with broken grammars and spellings variations. Nowadays, Internet memes come in almost every form of media, with modish formats continuously expanding. Initially, they work as a medium for humor to be shared, using cultural themes. However, they can also be manipulated to further political ideals, company promotions, and social media marketing.

\begin{center}
  \begin{figure}[h!]
  \makebox[\textwidth]{\includegraphics[width=5cm, height=5cm]{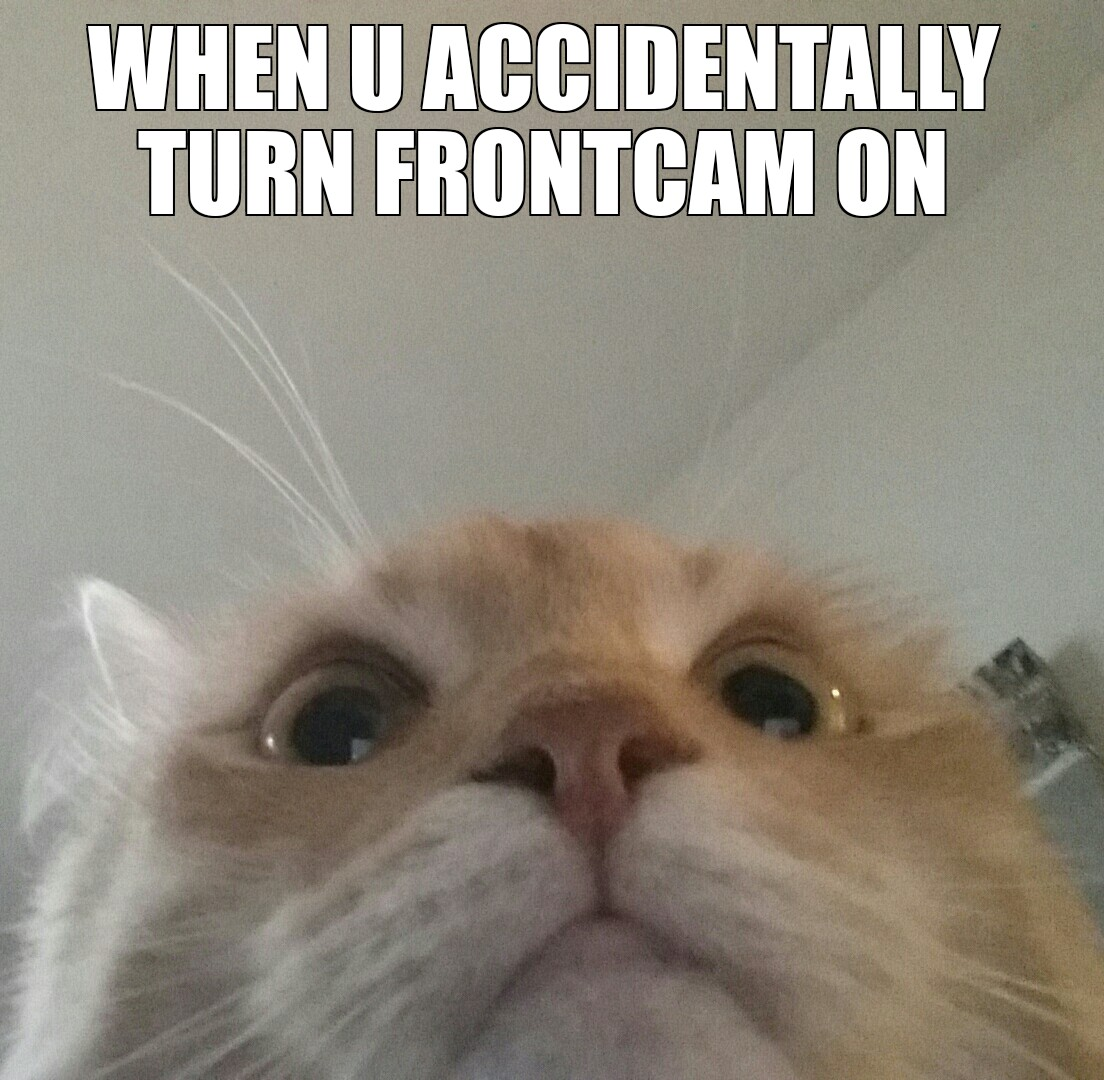}}
  \caption{Example of Cat Meme.}
  \end{figure}
\end{center}

In this paper, we present a multimodal approach for SemEval 2020: Task 8 on Memotion Analysis (Sharma et al., 2020). Task 8 consists of the following three subtasks:

\begin{enumerate}[label=(\Alph*)]
  \item Sentiment Classification: From the given input Internet meme extract the sentiment and classify it as a positive, negative, and neutral meme.
  \item Humor Classification:  In this sub-task, the system has to recognize the type of humor expressed in a given Internet meme. The humor classes are sarcastic, humorous, offensive, and motivational. A meme can have more than one class.
  \item Scales of Semantic Classes: In the third sub-task, quantify the extent to which a particular effect is being expressed. Details of such quantification are reported in Table 1.
\end{enumerate}

All our models are trained using only the corpus supplied by Memotion analysis organizers.  The evaluation metric for subtask \textbf{A} is macro F1 score and macro F1 score for each of the subtasks, and then average for subtask \textbf{B} and \textbf{C}. Primarily, we started experiments with traditional machine learning algorithms\footnote{https://scikit-learn.org/stable/} like Support Vector Machine (SVM) and Logistic Regression (LR). Based on the findings from them, we moved to deep learning models\footnote{https://keras.io/} like Long Short-Term Memory (LSTM), CNN in search of the better model. 

In the next section, we summarize the related work. Section 3 gives details about the dataset. We present the experimental setup, model description, and comparison with baselines \& other methods in Section 4. Results and Error analysis constitutes in Section 5. Finally, we conclude this paper in Section 6.

\section{Related Work}

Sentiment analysis was pioneered for text (Dmitry Davidov et al., 2010; Johan Bollen et al., 2011; Xia Hu et al., 2012) but image-based sentiment analysis has got much less attention compared to text-based sentiment analysis.  As Internet memes is a relatively new research topic, our research work is broadly related to studies on predicting the sentiment from the visual imagery in online reviews (Troung et al., 2017). Borth et al. (2013) introduced sentiment analysis on large scale visual sentiment ontology with SentiBank, a system extracting mid-level semantic attributes and from images.  To inspect the image posting behavior of social media users Chen et al. (2015) developed the Visual Emotional Latent Dirichlet Allocation (VELDA) model to capture the text-image correlation from multiple modalities: the text, visual and emotional perspective of the picture. And interestingly they found results that 66\% of users adding an image to their social media posts.

Recently, Anthony Hu et al. (2018) developed a multimodal sentiment analysis approach that combines text and vision to forecast the emotion word tags attached by users to their Tumblr posts. You et al. (2015) developed a robust Image Sentiment Analysis system by using a CNN on Flickr with domain transfer from Twitter for binary sentiment classification. Benet Sabat et al. (2019) worked on a challenge of automatically detect the hate speech in Internet memes by using visual information. Philipp Blandfort et al. (2019) developed a Multimodal Social Media system to study how public tweets with images posted by teens who mention gang violence on Twitter can be leveraged to automatically discover psycho-social factors and problems.

Few researchers have investigated to automate the Internet meme generation process, while a few others tried to extract its sentiment. To generate popular meme descriptions, William Yang Wang and Miaomiao Wen (2016) proposed a non-paranormal approach by combining visual and textual features. To produce the meme descriptions Peirson et al. (2018) presented an encoder-decoder meme generating system, consisting of a Google's pre-trained Inception-v3 network to generate an image embedding, followed by LSTM model with attention. In this paper, we refer to Wang and Hua’s (2014) method to combine textual and vision information, while scaling up the model using effective dropout regularization.

\section{Dataset}

We used the corpus provided by Task 8 in SemEval 2020. This task is named as \say{Memotion Analysis}. The dataset consists of Internet memes and corresponding text descriptions. For Task A, each meme is labeled into three sentiment classes: Positive, Neutral, or Negative, and for Task B and C, each meme is labeled with several sub-tasks: Humorous, Sarcastic, Offensive, and Motivational. Table 1 shows the detailed labeling of each task.

\begin{table}[h!]
\centering
\begin{adjustbox}{width=\textwidth}
\begin{tabular}{c cccc cccc}
\toprule
 & \multicolumn{4}{c}{ Humor Classification (Task B)} & \multicolumn{4}{c}{Semantic Classes (Task C)} \\
\cmidrule(lr){2-5} \cmidrule(lr){6-9}
Sentiment Labels (Task A)     & Humour   & Sarcastic  & Offensive & Motivational & Humour   & Sarcastic  & Offensive & Motivational \\
\midrule
Positive & Yes & Yes & Yes  & Yes  & Funny  & General & Offensive & Motivational \\
Neutral & No & No & No & No  & Very Funny  & Twisted Meaning & Slight & Not Motivational \\
Negative &  &  &   &   & Hilarious  & Very Twisted & Very Offensive &  \\
 &  &  &   &  & Not Funny  & Not Sarcastic & Hateful Offensive & \\
\bottomrule
\end{tabular}
\end{adjustbox}
\caption{\label{font-table} Labels of each task. }
\end{table}

Table 2 shows the distribution of each task labels in the dataset. Our dataset contained 6990 memes and corresponding descriptions for training and 914 memes for testing. Organizers did not provide any dataset for development so we split the training dataset into the train (85\%) and dev (15\%).

\begin{table}[h!]
\centering
\begin{adjustbox}{width=\textwidth}
\begin{tabular}{ccc}
\toprule
\textbf{TASKS}(in bracket each task labels) & \textbf{TRAIN}(in bracket each label counts) & \textbf{VALIDATION} \\
\midrule
\textbf{Task A} &  \\
(Positive-Neutral-Negative) & (4160-2200-630) & (564-279-71)  \\
&\\
\textbf{Task B} &   \\
(Humorous-Not Humorous) & (1650-5340) & (704-210)  \\
(Sarcastic-Not Sarcastic) & (394-5446) & (681-233) \\
(Offensive-Not Offensive) & (2713-4277) & (545-369)  \\
(Motivational-Not Motivational) & (4524-2466) & (589-325)  \\
&\\
\textbf{Task C} & \\
(Not Funny-Funny-Very Funny-Hilarious) & (1650-2452-2238-650) & (210-315-310-79) \\
(Not Sarcastic-General-Twisted Meaning-Very Twisted) & (1544-350-1546-394) & (233-444-195-42) \\
(Not offensive-Slightly offensive-Very Offensive-Hateful Offensive) & (2713-2592-1465-220) & (369-323-198-24) \\

\bottomrule
\end{tabular}
\end{adjustbox}
\caption{\label{font-table} Distribution of labels in each task. }
\end{table}

\subsection{Pre-processing}

While dealing with meme dataset, some of the major challenges faced were word or phrases with multiple spelling variations, short sentences with unclear grammatical structure, memes without any images and only text, URLs, HTML tags, and imbalanced dataset. To tackle some of the above issues we took the following pre-processing steps: 

\begin{enumerate}
  \item \textbf{Removal of URLs:} Text/Meme descriptions which are extracted from Internet memes, contain URLs (GrumpyCatPics.com, memegenerator.net, etc). We removed these URLs since it does not contribute towards our goal.
  
  \item \textbf{Removal of HTML Tags, Punctuation Marks, Digits, and Non-ASCII Glyphs:} All the HTML tags, punctuation marks, digits, and non-ASCII glyphs in a meme description are removed.
  
  \item \textbf{Handling Usernames and Hashtags:} Replaced the usernames with a \url{USER} tag. Pound (\#) sign removed from the hashtag and split into words based on digits and capital letters. Example: \(\text{\#10YearChallenge} \rightarrow \text{10 Year Challenge}\).
  
  \item \textbf{Handling of Word/Phrase Contractions:} Created a word/phrase contractions dictionary which contains around 250 words. By using this dictionary mapped the unusual words/phrases to proper English words. Examples: \(\text{gng} \rightarrow \text{going}\), \(\text{ASAP} \rightarrow \text{as soon as possible}\).  
  
  \item \textbf{Handling of Elongated Words:} By using regex converted the elongated words to standard English words. Examples: \(\text{Nooooo} \rightarrow \text{No}\), \(\text{suuuppperrr} \rightarrow \text{super}\). 
  
  \item \textbf{Handling of Imbalanced Datasets:} Data imbalance usually reflects the number of observations per class which is not equally distributed within the dataset. If we look at our dataset, we observe the same kind of problem.  To handle this problem we randomly duplicate samples in the minority class called \textbf{Random Oversampling} with the help of sklearn\footnote{https://github.com/scikit-learn-contrib/imbalanced-learn}.
  
\end{enumerate}
  
\section{Our Approach}

In this section, we present three different types of deep neural network architectures. We model each task in Memotion Analysis as a multi-class classification problem where given an Internet meme and text description, the model outputs probabilities of it belonging to output classes. The number of output classes will vary for each individual task. The full code of system architecture can be found on GitHub\footnote{https://github.com/SunilGundapu/Memotion-Analysis} .

\subsection{Bidirectional Long Short-Term Memory (BiLSTM) Network with Glove Embeddings}

BiLSTM is a recurrent neural architecture (Schuster and Paliwal, 1997) consists of forward and backward LSTM's. The forward LSTM reads the input text in forward order and uses the contextual information from the past. The backward LSTM reads the text description in the reverse order and preserves the contextual information from future. These two LSTM's generate two independent sequence output vectors. We obtain an output vector for each word by concatenating these forward and backward vectors. Figure 2 shows the architecture of BiLSTM with Glove embeddings.

\begin{center}
  \begin{figure}[h!]
  \makebox[\textwidth]{\includegraphics[width=11.5cm,height=6cm]{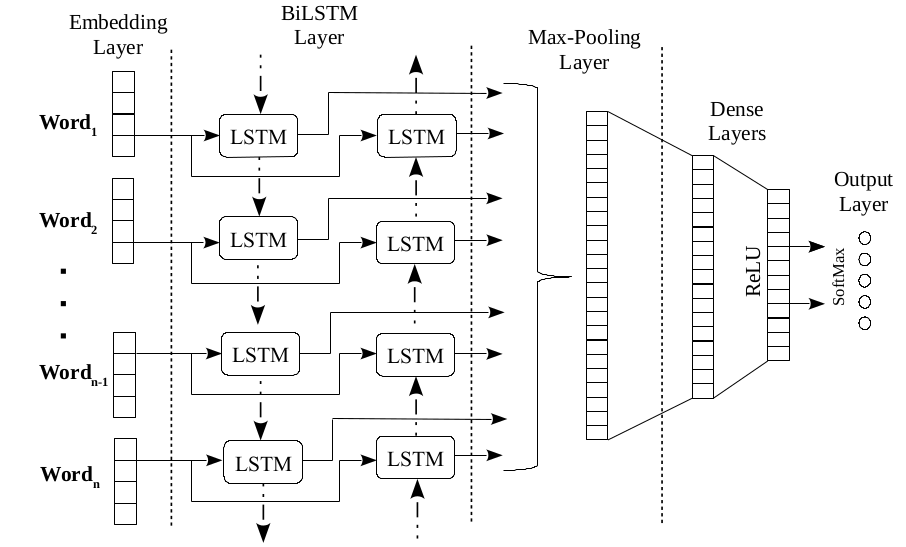}}
  \caption{BiLSTM network with Glove embeddings}
  \end{figure}
\end{center}

The input text description is tokenized and fed to the embedding layer. The embedding layer maps the input sequence to a matrix of shape \(\textbf{n} \times \textbf{d}\), \textbf{n} is the number of words in the text description, and \textbf{d} is the dimension of the Glove vector. Output matrix of an embedding layer used to feed the BiLSTM layer. A dropout of 0.2 was applied to the input of the BiLSTM layer and a dropout of 0.1 was used for the output of BiLSTM layer. After BiLSTM layer placed a global max-pooling layer, resulting in an output shape of \(\textbf{d} \times \textbf{1}\). This output is forwarded to two Dense layers with the activation of Rectifier Linear Unit (ReLU). Dense layer output was passed through a SoftMax layer having \textbf{m} units. The dimension of \textbf{m} depends on the number of classes in each task.

\subsection{Multimodal Neural Network (MNN) - I}

In this section, we demonstrate a multimodal neural network architecture for memotion analysis. In the previous architecture we use the only textual features but in this multimodal architecture join two different data modalities (text and image) for better results. 

\subsubsection{Image Embeddings}

Training a CNN model from scratch can be challenging as a huge amount of data is required and many different models have to be tried before achieving satisfying performances. To avoid this issue, we are using 42-layer deep learning pre-trained network called Inception-v3 that trained to recognize images through the ILSVRC-2012-CLS\footnote{http://image-net.org/challenges/LSVRC/2015/} image classification dataset. The third edition of Google Inception network stacks 11 inception modules where each module consists of convolutional filters with rectified linear units and pooling layers.  

\subsubsection{System Architecture}

Figure 3 presents a detailed description of the architecture. On the one side of architecture, the internet meme image, resized to (224,224) is run through the Inception-v3 network, the output is a vector of shape \(\textbf{2048} \times \textbf{1}\) called image embedding, that captures the content of the image. The image embedding is forward to a fully connected layer that converts the vector of shape \(\textbf{2048} \times \textbf{1}\) into  the vector of shape \(\textbf{128} \times \textbf{1}\).

\begin{center}
  \begin{figure}[h!]
  \makebox[\textwidth]{\includegraphics[width=11cm, height=7cm]{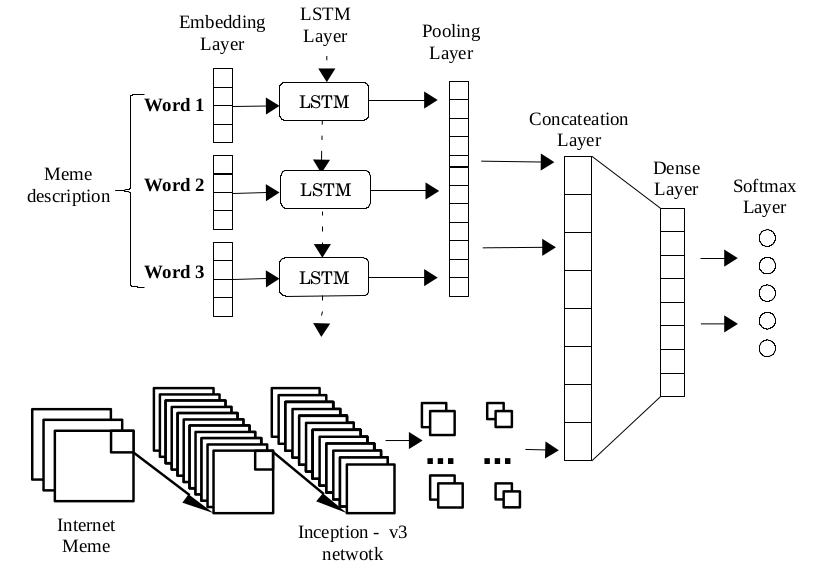}}
  \caption{Architecture of Multimodal Neural Network - I}
  \end{figure}
\end{center}

On the other side, the preprocessed meme description is tokenized into a sequence of words and pad with zeros so that each sequence has length \textbf{n} (=75). The word embedding layer maps the input sequence to a matrix of shape \(\textbf{n} \times \textbf{d}\), here \textbf{d} (=200) is the dimension of the Glove vector. The output matrix is feed to a LSTM layer, outputs a vector of shape \(\textbf{128} \times \textbf{1}\). 

Further, concatenation of both the textual and image modalities results in a vector of shape \(\textbf{256} \times \textbf{1}\). In the end, we have a dense layer with the activation of ReLU followed by a softmax layer. The output from the softmax layer is a vector of shape \(\textbf{m} \times \textbf{1}\). That refers to class probabilities for the \textbf{m} classes in a particular task. We used both PyTorch\footnote{https://pytorch.org/} and Keras libraries to build this model.

\subsection{Multimodal Neural Network - II}

We experiment this model with two word embedding layers and one image embedding layer. In two word embedding layers, one is Sentiment Specific Word Embeddings (SSWE), and the other is Glove embeddings. Gupta et al. (2017) proved that GloVe embeddings capture semantic information and SSWE (Tang  et  al.,2014) embeddings capture sentiment information in the continuous sequence of words.

\begin{center}
  \begin{figure}[h!]
  \makebox[\textwidth]{\includegraphics[width=\textwidth, height=7.5cm]{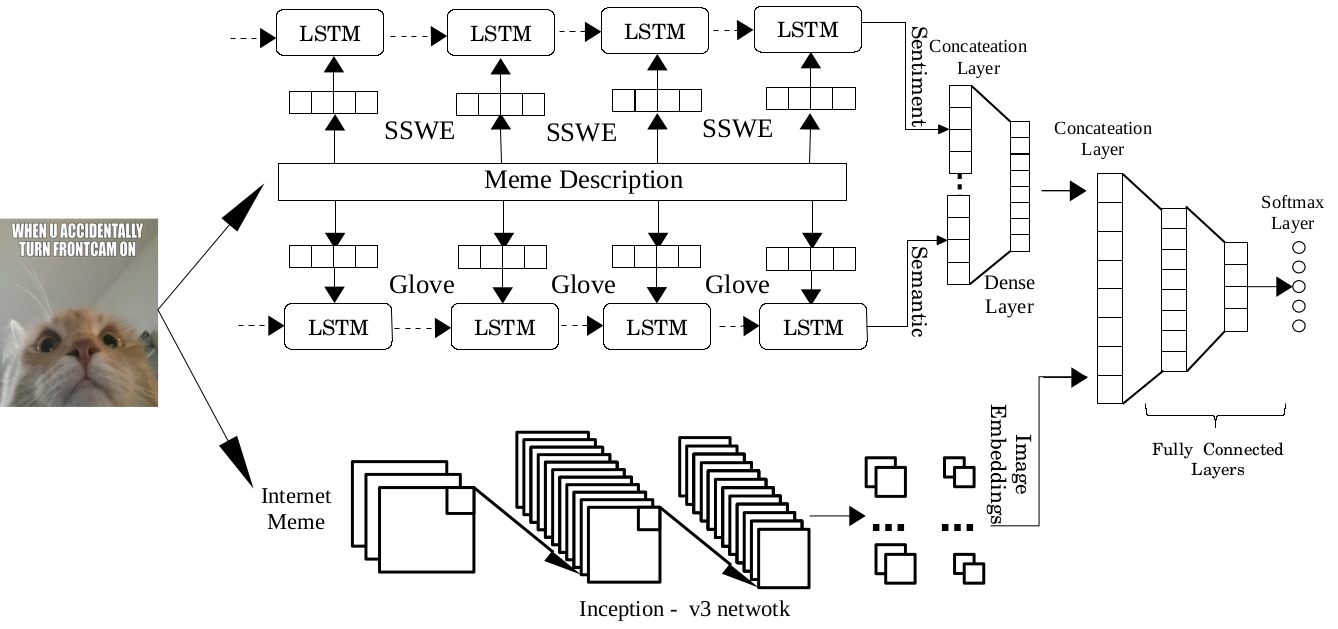}}
  \caption{Architecture of Multimodal Neural Network - II}
  \end{figure}
\end{center}

An overview of the system architecture can be found in Figure 4. As already said like our approach is built on two components (Word embeddings, Image embeddings). Initially, the input meme image is forwarded to the image embedding layer that generates image feature representations by using the Inception-v3 network. The output dimension of image embeddings is \(\textbf{2048} \times \textbf{1}\) is projected in a vector shape of \(\textbf{256} \times \textbf{1}\) by using a Dense layer. Then we passed the input as a meme description to two LSTM layers using two word embedding matrices with the shape of \(\textbf{n} \times \textbf{d}\). One layer uses a sentiment specific word embedding, whereas the other layer uses a Glove word embedding. From meme description, these two word embedding layers comprehend sentiment and semantic feature representation and observe sequential patterns. These two feature representations are concatenated and passed to a Dense layer with ReLU as an activation function. The output is a vector of dimension \(\textbf{256} \times \textbf{1}\).

The two modalities (image and text) outputted vectors are then concatenated and the output is a vector of shape \(\textbf{512} \times \textbf{1}\). The concatenated vector is passed through a fully connected network and then a softmax output layer to give the probability distribution over the different classes in a task.

\section{Results}

In Table 3, we show the performance and comparison of all the systems presented in Section 4. We observed that Multimodal Neural Network approaches perform better than benchmark models for all the three tasks. These MNN approaches capture the sentiment along with semantic information in the meme description and extracting the emotion in internet memes. 

By comparing the results of the three models, MNN-I outperforms for task A and MNN-II outperforms for tasks B and C in the official macro F1 score. Bi-directional LSTM with Glove vectors did not yield very good results. This system failed to capture the meaning of
polysemous word in different contexts and failed to handle short sentences.
The MNN systems performed very well on the noisy and imbalanced dataset and failed in some cases such as identifying sarcasm, extracting the emotion (when text covered the expression in memes), and identifying visual cue in the Internet memes.

\begin{table}[h!]
\centering
\begin{adjustbox}{width=\textwidth}
\begin{tabular}{c|cc|cc|cc}
\toprule
 & \multicolumn{2}{c}{\textbf{Task A}} & \multicolumn{2}{c}{\textbf{Task B}} &
 \multicolumn{2}{c}{\textbf{Task C}}\\
\cmidrule(lr){2-3} \cmidrule(lr){4-5} \cmidrule{6-7}
\textbf{Model}    & F1-Score (Macro)   & F1-Score (Micro)  & F1-Score (Macro)   & F1-Score (Micro) & F1-Score (Macro)   & F1-Score (Micro) \\
\midrule
\textbf{Baseline} & 0.2176 & 0.3077 & 0.5002  & 0.5686  & 0.2483  & 0.3328 \\
\textbf{BiLSTM} & 0.2984 & 0.3848 & 0.4713  & 0.5364  & 0.2991  & 0.3236 \\
\textbf{MNN - I} &\textbf{0.3391} & \textbf{0.4627} & 0.4944  & 0.5846  & 0.3074  & 0.3420 \\
\textbf{MNN - II} & 0.3261 & 0.3972 & \textbf{0.5014}  & \textbf{0.5896}  & \textbf{0.3123}  & \textbf{0.3489} \\
\bottomrule
\end{tabular}
\end{adjustbox}
\caption{\label{font-table} Results of different systems. }
\end{table}

To find the right set of hyper-parameters, we used the grid search and development dataset. By considering the dropout (Srivastava et al., 2014), we found the hyper-parameters like the number of LSTM layers, number of epochs, and learning rate. We used the GPU for training all our models.

\section{Conclusion}

In this paper, we developed a novel multimodal neural network method using deep learning techniques. Our model is constructed by concatenating the textual and image feature representations. We discovered that a combination of text and vision modalities give better predictions than single modalities (text or vision) for memotion analysis. Till now we handled challenges like the unstructured/elongated words, phrase and word contractions, multiple sentences, noisy data, imbalanced datasets, etc. In future work, we will concentrate on problems like short meme descriptions, free word ordering in sentences, more features to identify expressions in memes, sarcasm in descriptions, etc. We would like to explore more deep neural network architectures that are able to capture humor and sarcasm in Internet memes.


\end{document}